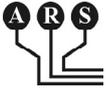
*Wang, C.; Chen, X.; Zhao, X. & Ju, S. / Design and Implementation of a General Decision-making Model in RoboCup Simulation, pp. 207 - 212, International Journal of Advanced Robotic Systems, Volume 1, Number 3 (2004), ISSN 1729-8806*# Design and Implementation of a General Decision-making Model in RoboCup Simulation

**Changda Wang**[1][2]; **Xianyi Chen**[1]; **Xibin Zhao**[1] & **Shiguang Ju**[1]

[1] School of Computer Science and Telecommunication Engineering, Jiangsu University, China

[2] School of Computer Science, Carleton University, Canada, K1S 5B6*Abstract: The study of the collaboration, coordination and negotiation among different agents in a multi-agent system (MAS) has always been the most challenging yet popular in the research of distributed artificial intelligence. In this paper, we will suggest for RoboCup simulation, a typical MAS, a general decision-making model, rather than define a different algorithm for each tactic (e.g. ball handling, pass, shoot and interception, etc.) in soccer games as most RoboCup simulation teams did. The general decision-making model is based on two critical factors in soccer games: the vertical distance to the goal line and the visual angle for the goalpost. We have used these two parameters to formalize the defensive and offensive decisions in RoboCup simulation and the results mentioned above had been applied in NOVAURO®, original name is UJDB, a RoboCup simulation team of Jiangsu University, whose decision-making model, compared with that of Tsinghua University, the world champion team in 2001, is a universal model and easier to be implemented.*
***Keywords***: *decision-making model, decision coupling, RoboCup simulation, agent.*## 1. Introduction

Agent, as an important new concept in computer science, has been applied in AI, DAI (distributed artificial intelligence) and Software Engineering, etc. It provides a new way to analyze, design and implement an open distributed system. In the field of artificial intelligence, agent is regarded as an autonomous object with lifecycle in certain circumstances, and a system consisting of multiple agents that interact with and interrelate to each other is called a multi-agent system or MAS. In recent years, RoboCup simulation as a kind of typical MAS, is the hotspot in AI and DAI. It has a special theoretical significance and a most promising future of application. In RoboCup simulation, each player can be looked as an agent [1], and any team, who wants to win the game, will not only need to consider collaboration and coordination within the team but also take into account such factors as the rival team and the environment.

This paper will mainly discuss the general decision-making model of NOVAURO®, a RoboCup simulation team of Jiangsu University. In the second section, the literature review of RoboCup is given. In the third section, the principle of the decision-model in NOVAURO® is expounded. In the fourth section, the basic offensive and defensive functions are established on the base of the study of real soccer games, and used to estimate the effectiveness of offensive and defensive actions of each player in the game. In the fifth and sixth sections, how to make the offensive and defensive decision based on the work mentioned above are specified respectively. In the seventh section, how to estimate the reliability of the communications between different players under noisy circumstances is discussed. In the eighth section, the universal character of NOVAURO® is illustrated through comparison with that of Tsinghua University, the world champion team in 2001. In the final section, the contribution of this paper and the future work of NOVAURO® are specified.

## 2. Literature Review

The RoboCup Federation is located in Switzerland and has about forty member countries. Since 1997, the Robot World Cup Soccer Games and Conferences, managed by the RoboCup Federation, are held every year. The ultimate goal of the RoboCup project is, by 2050, to develop a team of fully autonomous humanoid robots

207

that can win against the human world champion team in soccer games. Currently, RoboCup has five leagues, *i.e.* Simulation League, Small Size Robot League, Middle Size Robot League, Sony Legged Robot League and Humanoid League. The main focus of this paper is the Simulation League.

In RoboCup Simulation League, the game platform is SoccerServer issued by the RoboCup Federation. SoccerServer simulates a virtual game environment with two dimensions, the player controlled by Client can run and kick on the playground. Each team has 11 players and each player is an autonomous subject and can communicate with SoccerServer by UDP/IP [2]. The main functions of SoccerServer are to act as the referee as well as to accept and manage the request of players from each team, calculate the position and velocity of each object on the playground by mechanical theorem and send the aural and visual message to each player on the playground. The message each player gets from SoccerServer is accompanied by noise, and the player-to-player communication is only one-way and unreliable [3]. On the playground, balls and players have such attributes as size, position, velocity and acceleration, etc. Moreover, the player also has the attributes of direction and physical force.

Although the Deep Blue had beaten the human world champion in chess, the Robot still has a long way to go before the final success is achieved. The main differences between chess and soccer are showed in table 1[4].

|  | Chess | RoboCup |
| --- | --- | --- |
| Environment | Static | Dynamic |
| State Change | Turn taking | Real time |
| Info. accessibility | Complete | Incomplete |
| Sensor Readings | Symbolic | Non-symbolic |
| Control | Central | Distributed |

Table 1. Comparison of Chess and RoboCup

These differences show that each robot in RoboCup should be an autonomous subject that can collaborate with his teammates, and a RoboCup team should be a multi-subject system that allows inner cooperation. On the contrary, Deep Blue essentially is a program that can search 60 billion chess manuals in few minutes. Just for these reasons, to design RoboCup simulation team is a piece of work full of special techniques. To give different decision algorithms for different tactics is a popular method, and has been illustrated by all of the RoboCup simulation teams that had published their decision algorithms. For instance, Tsinghua University's team, the RoboCup simulation world champion team in 2001, used five different decision algorithms for ball handling, pass, shoot, interception and the movement of the players that do not control the ball [5]. Furthermore, no obvious relationships between these five different tactics can be found currently.

The experience of earlier edition of NOVAURO® shows that how to coordinate the relationships between these algorithms and how to chose the proper input parameters are the dilemma of such method mentioned above. For instance, NOVAURO® beat "11 monkeys", a RoboCup simulation team, with 9:0 in a test match, but we lost subsequently with 0:3 when just an input parameter had been changed! Nobody knew how to work out the optimum parameters, even after the math professors were consulted. And this is why most of RoboCup simulation teams are ready to publish their decision algorithm but their input parameters, the most valuable wealth derived from repetitious interior test.

### 3. The Principle of Decision-making Model Building

*3.1 Make Decisions under the Environment Provided by RoboCup SoccerServer*

RoboCup SoccerServer provides a simulation environment similar to a real soccer game, but some differences do exist, and all of the decisions must be made under such a given environment. For instance, the field provided by RoboCup SoccerServer is a 2-dimension surface, so any study of the ball flight in air is meaningless. For more details about RoboCup simulation, please visit its official website at http://www.robocup.org.

*3.2 Use Mature Tactics in Soccer Games to Guide Decision-making Model Design*

Via "Soccer Coach Manual" and investigation of some soccer players, we select the vertical distance to the rivals' goal line and our shooter's visual angle for the goalpost as fundamental parameters to design the decision-making model. We also design the "vertical pass incline arrive" and "incline pass vertical arrive" as the scoring tactics (Fig. 1, 2).

*3.3 Decision Formalized and Task Decomposed*

Because the match is held in an environment provided by RoboCup SoccerServer, so all of the decision-making models must be formalized and ultimately executed by computer.

As a principle, a complex task is usually decomposed into some smaller and easier ones in a project. Similarly, almost all of the popular RoboCup simulation teams adopt this hierarchical tactics [6].

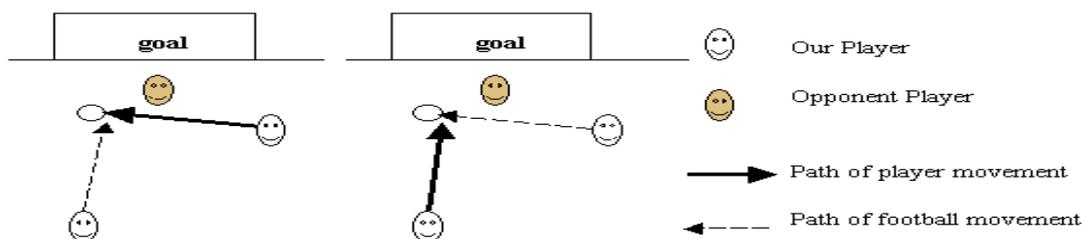

Fig. 1.  Vertical pass inclined arrive.     Fig. 2.  Incline pass vertical arrive.

## 4. Shooting Evaluation Function and Defensive Evaluation Function

*4.1 Shooting Evaluation Function*
*1) Hypothesis of Shoot Success Efficiency*
In our research, we find the factors that affect shooting success are as follow: (Fig.3)

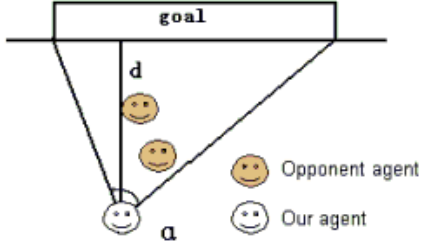

Fig. 3. Factors influencing shooting success.

(1) Vertical distance from the shooting agent to the opponent goal line, marked as *d*;
(2) Visual angle from the shooting agent to the opponent goalpost, marked as α;
(3) Shooting agent's ability, marked as *f*;
(4) The degree of the opponent agents' interference with our shooting agent, marked as ξ;

*2) Basic Hypotheses*
(1) If the ball is placed on the opponent goal line and no opponent agents interference exists, i.e. $d=0$, $α=π$, $ξ=0$, scoring will be a certainty (Fig. 4). *(hypothesis1)*
*3) Shooting Success Evaluation Function*
(2) If the ball is situated on the opponents' extended goal line, i.e. $d =0$, $α=0$, scoring is impossible (Fig. 5). *(hypothesis 2)* [Though at the real soccer game this situation also can be scored, but in Robocup it's impossible.]
(3) Shooting success is in direct proportion with α and *f*, where α is the visual angle from the shooting agent to the opponents' goalpost and *f* is the ability of the shooting agent. *(hypothesis 3)*
(4) Shoot success is in inverse proportion with $d^2$ and ξ, where *d* is the vertical distance from the shooting agent to the opponents' goal line and ξ is the degree of the opponent agents interference with the shooting agent. *(hypothesis 4)*.
Where α is the visual angle from the shooting agent to the opponents' goalpost, *f* is the ability of the shooting agent, *d* is the vertical distance from the shooting agent to the opponents' goal line and ξ is the degree of the opponent agents interference with shoot agent.

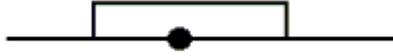

Fig. 4. Football placed on goal line.

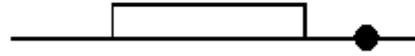

Fig. 5. Football placed on extend goal line.

**Model 1**

$$Shooting\_Success\ (d,\alpha,f,\xi) = \{\frac{1}{(1+d)^2}\cdot\frac{\alpha}{\pi}+\frac{f\cdot\alpha}{f\max\cdot(1+d)\cdot\pi}\cdot[1-\frac{1}{(1+d)^2}\cdot\frac{\alpha}{\pi}]\}\cdot\frac{1}{1+\xi} \quad (1)$$

*4) Properties of Shooting_Success Function*
We can easily calculate the following result: ("≡" is identity sign)

- Shooting_Success(0,π,f,0) ≡ 1  (satisfy *hypothesis1* in this paper);
- Shooting_Success(0,0,f,ξ) ≡ 0  (satisfy *hypothesis 2* in this paper);

We can prove that *Shooting_Success* function also meets the requirement of *hypothesis 3* and *hypothesis 4* in this paper by mathematical method, but we will not give the full demonstration due to limited space. The diagrams of *Shooting_Success* function drawn by Mathcad 2000 show clearly that as *d* and ξ increase, the value of *Shoot_Success* function will decrease, while as α and *f* increase, the value of *Shoot_Success* function will also increase. The value region of *Shooting_Success* function falls on [0, 1].

*5) Preferences of Shooting_Success Function*
Values of *d* and *f* can be calculated by functions provided by RoboCup SoccerServer. α can be calculated by the following formula:

$$\alpha = \arccos\frac{(x1-x)\cdot(x2-x)+(y1-y)\cdot(y2-y)}{\sqrt{(x1-x)^2+(y1-y)^2}\cdot\sqrt{(x2-x)^2+(y2-y)^2}} \quad (2)$$

Where (x, y) is the plotting of the shooting agent, (x1, y1) is the plotting of the opponents' left goalpost, (x2, y2) is the plotting of the opponents' right goalpost.
Values of ξ are discrete and determined by the number of the opponent agents and their distance to the shooting agent within his visual angle. Given ξ ∈ {0, 1, 2, 3, …} and as the value increase, the interference to the shooting agent will duly increase.

*4.2 Defensive Evaluation Function*
Defensive evaluation function is derived from *Shooting_Success* function, because the defensive RoboCup simulation team believes:
Firstly, it should prevent the opponent agent who controls the ball from going ahead to its goal line in the vertical direction, because *d* has a higher right in *Shoot_Success (d, α, f, ξ)*.
Secondly, it should let the opponent shooting agent have the smallest visual angle for its goalpost.

**Model 2**

$$\text{Defensive}(d, \alpha) = \frac{1}{(1+d)^2} \cdot \frac{\alpha}{\pi} \quad (3)$$

Where *d* is the vertical distance from the defensive location to its goal line and α is the visual angle from the defensive location to its goalpost.

*1) Properties of Defensive Evaluation Function*
(1) For α ∈ [0,π], *d*≥0,
   so the range of Defensive(d, α) falls on [0, 1].

(2) For α≥0, *d*≥0,
   so $\frac{\partial \text{Defensive}(d, \alpha)}{\partial d} = -\frac{2\alpha}{\pi} \cdot \frac{1}{(1+d)^3} \leq 0$, *i.e.* as *d* increases, the value of *Defensive(d, α)* function decreases. So when the visual angle for the goalpost is constant, the aim of defense is to let the opponent agent have the longest vertical distance to the goal line.

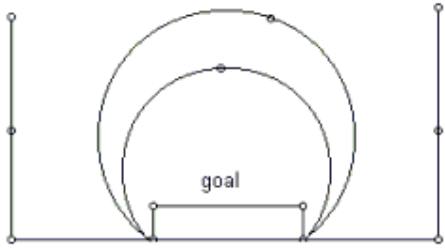

Fig. 6. Isoline having the same visual angle

(3) For α≥0, d≥0,
   so $\frac{\partial \text{Defensive}(d, \alpha)}{\partial \alpha} = \frac{1}{(1+d)^2} \cdot \frac{1}{\pi} > 0$ *i.e.* as α increases, the value of Defensive(d, α) function increases. So when the vertical distance to the goal line is constant, the aim of defense is to let the opponent agent have the smallest visual angle.

Because circle angles on the same arch are equal, all the locations have the same visual angle to the goalpost on the arch whose bowstring falls on the goal line (Fig. 6)

## 5. Offensive Decision

*5.1 Offensive Decisions of Individual Agents*
This kind of decisions requires the agent that controls the ball to calculate its *Shooting_Success* value in every RoboCup SoccerServer time-cycle. If its value is greater than the shooting critical value we have given in advance, then the agent will execute the shooting action. *e.g.* if the shooting critical value is 0.6, it means the chance of scoring is 60%. Else it will intercept the *Shooingt_Success* values sent via "*say*", a command provided by RoboCup SoccerServer, by its teammates who are free, then the agent will pass the ball to a teammate agent who has a higher *Shooting_Success* value. If no such a teammate agent exists, it will move to a new location with a higher *Shooting_Success* value. Repeat the above process until it executes the shoot action or its ball is intercepted by an opponent agent.

We only require those agents who can see an opponent agent in their visual angle to use the "*say*" command to broadcast his *Shooting_Success* value, because agents who can't see an opponent agent in their visual angle may be offside already.

*5.2 Collective Offensive Decision of Agents*
All the agents within a RoboCup simulation team, as a whole, do not completely rely on communications, just like a well-trained team of humans, no or little communication is required to let teammate agents understand each other. We call this as "decision coupling". The principle of this "decision coupling" can be presented as when agent_1 calls one of the branches in his decision tree, and agent_2, though not having the whole information, can select an exact branch in his decision tree to achieve the collaboration with agent_1 relying on the experience of collaboration with agent_1 in previous training. Fig. 7 illustrates {A, B, D}, the branch of agent_ 1's decision tree, coupling with {a, c, f}, the branch of agent_2's decision tree. For instance, our tactics of vertical pass incline arrive and incline pass vertical arrive at a point before the opponent goal (Fig. 1, 2) rely on the privities between the passing agent and the shooting agent. The passing agent passes the ball to a certain location and the shooting agent runs to this location and then executes the shooting action. It must be understood that the communication between two agents does not include the plotting of that location, but merely a secret signal, which means the information is not full. Different secret signals mean the selection of different branches of the decision tree, and different locations of passing the ball and arrival.

There are at least two advantages in this design: Firstly, the use of secret signal will baffle the opponents in their effort to decode the communications among our teammate agents. Secondly, in the noisy game environment provided by RoboCup SoccerServer, decrease in the volume of communications means increase in the reliability of communication.

## 6. Defensive Decision

*6.1 Formation*
As in a real soccer game, we design the "formation" concept for the agents of RoboCup simulation. Faced with different opponents, our agents will adopt different formations, such as 4-4-2 (Fig. 8) or 4-3-3, etc. The formation is implemented by assigning different offensive and defensive action areas for each agent on the field. Statistics show that formation is more effective in the defense than in the offensive.

It's needs to be understood that the partition of each agent's action area is not rigid but flexible. Considering the requirement of collaboration, we permit each agent to act in his 8-connection area (Fig. 9). In such an environment, we can see midfielder agents have the biggest action area and the forward agents and back agents have smaller action area correspondingly. This is consistent with real soccer match that a midfielder should connect its forwardline and backline.

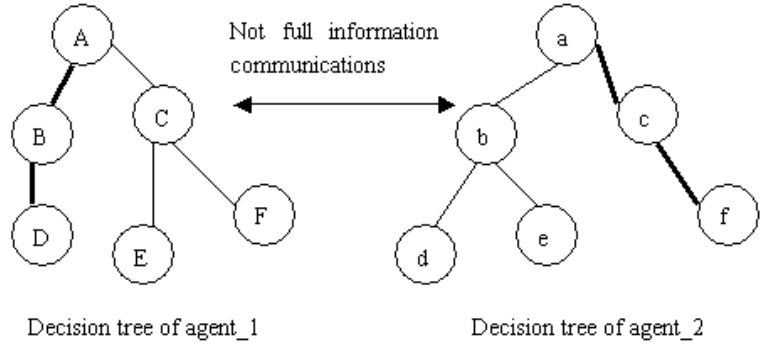

Fig. 7. Decision coupling trees.

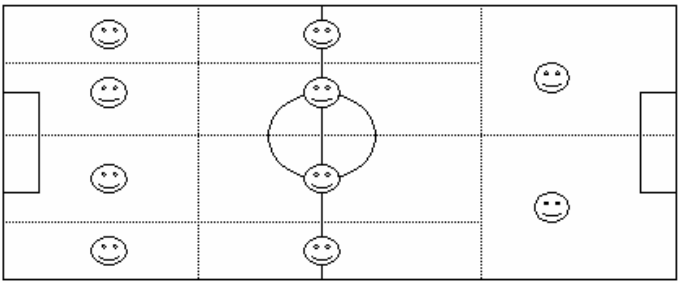

Fig. 8. Partition area of 4-4-2 formation.   Fig. 9. 8-connection area of F.

*6.2 The Movement of the Defensive Agents*
Firstly, if an opponent agent controls ball in his main defensive area or in his assistant defensive area, the defensive agent will trace the ball and execute the interception action.

Secondly, if the above-mentioned scenario does not happen and the defensive agent can be sure that the opponent is attacking now, it will select the grads of *defensive* function to move, *i.e.* in the following direction of movement:

$$Direction = gradDefensive(d,\alpha)$$
$$= (-\frac{2\alpha}{\pi} \cdot \frac{1}{(1+d)^3}, \frac{1}{(1+d)^2} \cdot \frac{1}{\pi}) \qquad (4)$$

In accord with field theory, the direction of grads is the direction where the value of *Defensive(d, α)* function increases most quickly, having nothing to do with the plotting system.

The use of grads may lead to part extremum, and this is why we do not use direction of grads as offensive agents' direction of movement. Suppose the defensive agent is trapped in part extremum, if the scenario as mentioned above emerges, then the agent will jump out of part extremum.

### 7. Reliability of Communication

Since the game environment offered by RoboCup SoccerServer is a noisy one, the vision and hearing of the agents in the field are all interfered. Generally speaking, the intensity of interference is in direct proportion with the distance between communication objects. When the distance is 100.0, the maximum of noise is up to 10.0, but when the distance is 10.0, the noise is less than 1.0. [5]

So an agent must evaluate the reliability of his communications.
**Model 3**

$$Believe(d) = \frac{1}{d^2} \qquad (5)$$

Where *d* is the distance between communication agents. If the value of *Believe(d)* function is less than a certain value given in advance, then we believe this communication is trustless, *i.e.* the message received under such circumstance can not be used as a decision bedrock.

### 8. Algorithm of NOVAURO® vs. Tsinghua University's RoboCup Simulation Team

Tsinghua University's RoboCup simulation team is the world champion in 2001. The following table (Table.2) shows the differences of decision algorithms between NOVAURO® and Tsinghua's team in some ta ctics.

These differences show that almost all of the decision algorithms of NOVAURO® for each tactics in soccer are based on the calculating results of *Shoot_Success(d,α, f,ξ)* function and its derived function *Defensive(d α)*, whereas the Tsinghua's team gives difference algorithms for different tactics. The decision model of NOVAURO® shows that different tactics should be interrelated in a soccer match but not disrupted, and it is a general one in a sense. Because the key parameters used by the decision algorithm of NOVAURO®, such as *d, α, f* etc, are derived from real time soccer matches other than selected by man and do not need repeated tests, such a RoboCup simulation team is easier to be implemented and cost less. With a tiny investment of 20,000 Chinese Yuan (about 2,300 U.S. dollars) into the software design of NOVAURO®, its latest achievement is won the competency for the final stage game with win 9, tie 2 and lost 3 in CRC2003 (Chinese Robot Competition). It's an amazing achievement for such a puny investment, only about 1/5 of the monetary input of other RoboCup Simulation teams in mainland China.

| Basic tactics | Algorithm of NOVAURO® | Algorithm of Tsinghua University's RoboCup simulation team[5] |
|---|---|---|
| Ball handling | Based on calculated result of *Shooting_Success(d, α, f, ξ)* | There are 8 directions for choice in range of $2\pi$ |
| Pass | Based on calculated result of *Shooting_Success(d, α, f, ξ)* | Select a fixed pass path from a certain aggregate |
| Shoot | Based on calculated result of *Shooting_Success(d, α, f, ξ)* | Shooting decision requires 3 degree interval scan of the angle from the ball to the goalpost; use its interception networks to evaluate the probability of the ball being intercepted by the opponents, and then select the shooting path. |
| Interception | By calculating the path of the ball that the opponent passes and intercepts it | Based on BP algorithm of neural network. |
| The movement of agents that do not control the ball | Based on calculated result of *Defensive(d α)* | Mainly consider the sensitivity of different locations in the field and the density of agents. Use the mutually exclusive function to make sure that multiple agents can't emerge in same area synchronously. |

Table 2. Contrast of Basic Tactics.

## 9. Conclusion

The main contribution of this paper is that we provide a general decision model for RoboCup simulation rather than select divided and unrelated algorithms for different tactics. We believe that the world is created by God according to fixed rules, and thus so perfect. Accordingly, we derived our defensive function by analyzing the offensive function. And the calculating result of these functions show that the location on the playground that has both higher offensive value also has higher defensive value, verifying an ancient military philosophy of China, *viz.* the antagonist's fatal point is our winning point – the antagonist's best location for offensive is our best location for defensive or vice versa.

Actually, the general decision-making model specified in this paper can be applied not only in RoboCup simulation but also in other RoboCup Leagues, *e.g.* Small Size Robot League, Middle Size Robot League, Sony Legged Robot League and Humanoid League, because any soccer robot need the software to control how it works. So this universal decision algorithm is valuable for any robot soccer games. We notice that further research need to be done on the following problems of this general decision-making model in the future development of NOVAURO®. It's also the most challenging work to be encountered. Firstly, how to distribute the strength of agents.In a RoboCup simulation match, each agent in the field has a certain value of strength. By restriction of the strength value, RoboCup SoccerServer prevents agents from running at its maximal speed all the time. Rational distribution of the strength of agents in the match will improve the battling might of the whole team. Secondly, interception and pass.Nowadays, all of the famous RoboCup simulation teams use C 4.5 to build decision tree or neural network to implement its interception and pass actions. Implementation of such an algorithm need huge amount of real case training, and hence is inefficient. Finally, in the long turn, our research will concentrate on dynamic task decomposition of the collaboration of agents. The existing algorithm of NOVAURO® is a rigid hierarchy model defined in advance. We hope the collaboration of agents could be a dynamic hierarchy in light with the task, *i.e.* a flexible hierarchy. The dynamic hierarchy will be more intelligent and adaptable.

**Acknowledgment**
Some ideas of this paper are derived from the discussion inside the development team of NOVAURO®. We want to thank Mr. Zhoumin Zheng, Mr. Ke fan, Mr. Wei Chu and Mr. Sheng Huang, who have joined the numerous discussions. We also want to thank the authority of JiangSu University, for the achievement of NOVAURO®, they had promised that more fund will invest in our group in 2004.